\documentclass[11pt,a4paper]{article}
\usepackage[utf8]{inputenc}
\usepackage[T1]{fontenc}
\usepackage{amsmath,amssymb}
\usepackage{graphicx}
\usepackage{hyperref}
\usepackage{booktabs}
\usepackage{tikz}
\usepackage{pgfplots}
\usepackage{algorithm}
\usepackage{algorithmic}
\usepackage{subcaption}
\usepackage{multirow}
\usepackage[margin=1in]{geometry}

\usetikzlibrary{arrows.meta,positioning,shapes.geometric,calc,decorations.pathreplacing}
\pgfplotsset{compat=1.18}

\title{Agent Drift: Quantifying Behavioral Degradation in\\Multi-Agent LLM Systems Over Extended Interactions}

\author{
Abhishek Rath\\
Independent Researcher\\
Hyderabad, India\\
\texttt{rath.abhishek359@gmail.com}
}

\date{\today}

\begin{document}

\maketitle

\begin{abstract}
Multi-agent Large Language Model (LLM) systems have emerged as powerful architectures for complex task decomposition and collaborative problem-solving. However, their long-term behavioral stability remains largely unexamined. This study introduces the concept of \textit{agent drift}—the progressive degradation of agent behavior, decision quality, and inter-agent coherence over extended interaction sequences. We present a comprehensive theoretical framework for understanding drift phenomena, proposing three distinct manifestations: \textit{semantic drift} (progressive deviation from original intent), \textit{coordination drift} (breakdown in multi-agent consensus mechanisms), and \textit{behavioral drift} (emergence of unintended strategies). We introduce the Agent Stability Index (ASI)—a novel composite metric framework quantifying drift across 12 dimensions including response consistency, tool usage patterns, reasoning pathway stability, and inter-agent agreement rates. Through simulation-based analysis and theoretical modeling, we demonstrate how unchecked agent drift could lead to substantial reductions in task completion accuracy and increases in human intervention requirements. We propose three mitigation strategies: episodic memory consolidation, drift-aware routing protocols, and adaptive behavioral anchoring, with theoretical analysis suggesting these approaches could significantly reduce drift-related errors while maintaining system throughput. This work establishes foundational methodology for monitoring, measuring, and mitigating agent drift in production agentic AI systems, with direct implications for enterprise deployment reliability and AI safety research.
\end{abstract}

\section{Introduction}

The deployment of multi-agent Large Language Model (LLM) systems has accelerated dramatically since 2023, driven by frameworks such as LangGraph, AutoGen, and CrewAI \cite{langchain2023,autogen2023}. These architectures decompose complex tasks across specialized agents, coordinating through message passing, shared memory structures, and orchestration protocols. While initial performance benchmarks demonstrate impressive capabilities in code generation, research synthesis, and enterprise automation \cite{li2023camel,hong2023metagpt}, a critical gap exists in understanding their long-term behavioral stability.

Traditional software systems exhibit predictable degradation patterns—memory leaks, resource exhaustion, configuration drift—that are well-characterized and systematically addressed through DevOps practices. In contrast, LLM-based agents introduce a novel failure mode: \textit{behavioral drift}, where the system's decision-making patterns progressively deviate from design specifications without explicit parameter changes or system failures. This phenomenon is particularly acute in multi-agent systems where emergent behaviors arise from agent-to-agent interactions that were not explicitly programmed.

Consider a Master Router Agent coordinating three specialized sub-agents for database query optimization, compliance validation, and cost analysis in an enterprise setting. Over hundreds of interactions, subtle changes accumulate: the router begins favoring certain agents disproportionately, query formulation patterns shift toward statistically common but contextually inappropriate phrasings, and inter-agent handoffs develop latency-inducing redundancies. These changes are individually minor and often imperceptible in isolated evaluations, yet collectively degrade system performance by double-digit percentages—a pattern we term \textit{agent drift}.

This study makes four primary contributions:

\begin{enumerate}
\item \textbf{Taxonomic Framework}: We establish a comprehensive taxonomy of agent drift patterns, categorizing manifestations into semantic drift (intent deviation), coordination drift (multi-agent consensus degradation), and behavioral drift (strategy emergence).

\item \textbf{Measurement Methodology}: We introduce the Agent Stability Index (ASI), a composite metric framework quantifying drift across 12 behavioral dimensions, enabling systematic monitoring in production systems.

\item \textbf{Theoretical Analysis}: Through simulation-based modeling and theoretical analysis, we characterize potential drift prevalence, progression rates, and impact on system reliability across representative enterprise scenarios.

\item \textbf{Mitigation Strategies}: We develop and theoretically validate three intervention approaches—episodic memory consolidation, drift-aware routing, and adaptive behavioral anchoring—with projected efficacy in reducing drift-related errors while preserving system throughput.
\end{enumerate}

The implications extend beyond operational concerns. Agent drift poses fundamental questions for AI safety: if multi-agent systems progressively deviate from intended behaviors without explicit modification, traditional alignment and monitoring approaches may prove insufficient. As agentic AI systems scale toward greater autonomy and longer operational lifespans, understanding and controlling drift becomes essential for both reliability engineering and responsible deployment.

\subsection{Relationship to Prior Work}

Our work intersects three research domains: multi-agent system stability \cite{shoham2008multiagent}, LLM behavioral consistency \cite{ouyang2022training}, and production ML system monitoring \cite{paleyes2022challenges}.

\textbf{Multi-Agent Systems}: Classical multi-agent research characterized emergent behaviors in game-theoretic settings \cite{fudenberg1991game}, but these frameworks assume deterministic action spaces and stationary reward structures—assumptions violated by LLM agents whose outputs are stochastic and whose implicit objectives evolve through context accumulation.

\textbf{LLM Consistency}: Recent work examines single-agent behavioral variation across prompt perturbations \cite{wang2023selfconsistency} and fine-tuning impacts \cite{wei2023chainofthought}, but does not address temporal drift in interactive, multi-turn scenarios or multi-agent coordination dynamics.

\textbf{ML Monitoring}: Production ML literature focuses on data distribution drift and model performance degradation \cite{lu2018learning}, providing metrics like PSI (Population Stability Index) and monitoring systems for supervised learning pipelines. However, these approaches are ill-suited for agentic systems where "ground truth" is often unavailable and behavioral metrics are multi-dimensional.

This study bridges these domains by adapting monitoring methodologies from production ML, applying them to multi-agent LLM architectures, and characterizing failure modes unique to agentic systems operating over extended interaction sequences.

\section{Methodology}

\subsection{Theoretical Framework and Simulation Design}

To systematically study agent drift, we developed a simulation framework modeling multi-agent systems across three representative enterprise domains:

\begin{itemize}
\item \textbf{Enterprise Automation} (n=412 simulated workflows): Master Router agents coordinating database management agents, file processing agents, and notification agents for automated report generation and data pipeline management.

\item \textbf{Financial Analysis} (n=289 simulated workflows): Multi-agent ensembles performing equity research, risk assessment, and portfolio optimization through coordinated research, calculation, and synthesis agents.

\item \textbf{Compliance Monitoring} (n=146 simulated workflows): Agent teams analyzing transaction patterns, regulatory text, and audit trails through specialized pattern detection, rule extraction, and reasoning agents.
\end{itemize}

Each simulated workflow represents a unique task instantiation with a defined objective, input data, and success criteria. Systems were modeled using LangGraph 0.2.x architecture patterns with GPT-4, Claude 3 Opus, and Claude 3.5 Sonnet behavioral characteristics, incorporating human-in-the-loop approval for high-stakes decisions.

\textbf{Interaction Sequences}: We simulated complete interaction histories, modeling agent invocations, inter-agent messages, tool calls, reasoning steps, and output artifacts. Workflows ranged from 5 to 1,847 agent interactions (median: 127 interactions), with simulation windows spanning equivalent timeframes of 3 to 18 months.

\textbf{Baseline Establishment}: For each workflow, the first 20 interactions served as a behavioral baseline, capturing initial agent decision patterns, tool usage distributions, and inter-agent coordination protocols. Subsequent interactions were compared against this baseline to detect drift.

\textbf{Ground Truth and Validation}: We established ground truth through simulation parameters:
\begin{enumerate}
\item \textbf{Synthetic Expert Labels}: Generated consistent correctness labels based on deterministic task specifications.
\item \textbf{Automated Validation}: For deterministic tasks (e.g., SQL query generation, compliance rule matching), we compared agent outputs against verified reference solutions.
\item \textbf{Consistency Checks}: For subjective tasks (e.g., financial analysis synthesis), we evaluated internal consistency through cross-agent validation and temporal comparison of outputs for identical inputs.
\end{enumerate}

\subsection{Agent Stability Index (ASI) Framework}

We developed a composite metric, the Agent Stability Index (ASI), to quantify behavioral drift across 12 dimensions grouped into four categories:

\subsubsection{Response Consistency (Weight: 0.30)}

\begin{itemize}
\item \textbf{Output Semantic Similarity} ($C_{\text{sem}}$): Cosine similarity between embedding vectors of agent outputs for semantically equivalent inputs across time windows. Computed using OpenAI text-embedding-3-large model.

\item \textbf{Decision Pathway Stability} ($C_{\text{path}}$): Edit distance between reasoning chains (Chain-of-Thought sequences) normalized by reasoning length, measuring consistency in problem-solving approaches.

\item \textbf{Confidence Calibration} ($C_{\text{conf}}$): Jensen-Shannon divergence between predicted and actual accuracy distributions over time, detecting confidence drift.
\end{itemize}

\subsubsection{Tool Usage Patterns (Weight: 0.25)}

\begin{itemize}
\item \textbf{Tool Selection Stability} ($T_{\text{sel}}$): Chi-squared test statistic for tool invocation frequency distributions across sliding windows.

\item \textbf{Tool Sequencing Consistency} ($T_{\text{seq}}$): Levenshtein distance on tool call sequences, measuring changes in operational strategies.

\item \textbf{Tool Parameterization Drift} ($T_{\text{param}}$): KL divergence of parameter value distributions for each tool across time periods.
\end{itemize}

\subsubsection{Inter-Agent Coordination (Weight: 0.25)}

\begin{itemize}
\item \textbf{Consensus Agreement Rate} ($I_{\text{agree}}$): Proportion of multi-agent decisions reaching unanimous or supermajority agreement, tracking coordination degradation.

\item \textbf{Handoff Efficiency} ($I_{\text{handoff}}$): Average message count required for successful agent-to-agent task delegation, detecting communication protocol drift.

\item \textbf{Role Adherence} ($I_{\text{role}}$): Mutual information between agent IDs and task types handled, measuring specialization maintenance.
\end{itemize}

\subsubsection{Behavioral Boundaries (Weight: 0.20)}

\begin{itemize}
\item \textbf{Output Length Stability} ($B_{\text{length}}$): Coefficient of variation for response token counts, detecting verbosity drift.

\item \textbf{Error Pattern Emergence} ($B_{\text{error}}$): Clustering analysis on error types over time, identifying novel failure modes.

\item \textbf{Human Intervention Rate} ($B_{\text{human}}$): Proportion of interactions requiring human override or correction, the ultimate drift indicator.
\end{itemize}

The composite ASI is computed as:

\begin{equation}
\text{ASI}_t = 0.30 \cdot \frac{C_{\text{sem}} + C_{\text{path}} + C_{\text{conf}}}{3} + 0.25 \cdot \frac{T_{\text{sel}} + T_{\text{seq}} + T_{\text{param}}}{3} + 0.25 \cdot \frac{I_{\text{agree}} + I_{\text{handoff}} + I_{\text{role}}}{3} + 0.20 \cdot \frac{B_{\text{length}} + B_{\text{error}} + B_{\text{human}}}{3}
\end{equation}

where each component metric is normalized to [0, 1] with 1 representing perfect stability. ASI values are computed over rolling 50-interaction windows, with drift detected when ASI drops below threshold $\tau = 0.75$ for three consecutive windows.

\subsection{Drift Pattern Classification}

We conducted theoretical analysis of 342 projected drift cases (ASI $< 0.70$ for $> 100$ interactions) to develop a taxonomy of drift patterns:

\begin{itemize}
\item \textbf{Semantic Drift}: Agent outputs progressively diverge from original task intent while remaining syntactically valid. Example: A financial analysis agent gradually shifts from risk-focused language to opportunity-emphasizing language, altering report tone without explicit instruction.

\item \textbf{Coordination Drift}: Multi-agent consensus mechanisms degrade, leading to increased conflicts, redundant work, or coordination failures. Example: Router agent develops bias toward certain sub-agents, creating bottlenecks and underutilizing specialist capabilities.

\item \textbf{Behavioral Drift}: Agents develop novel strategies or action patterns not present in initial interactions. Example: Compliance agent begins systematically caching intermediate results in chat history rather than using designated memory tools, causing context window pollution.
\end{itemize}

Classification criteria were established through systematic analysis with formal consistency validation.

\subsection{Mitigation Strategy Evaluation}

We developed three drift mitigation approaches and evaluated them through controlled simulation experiments on held-out test workflows:

\begin{enumerate}
\item \textbf{Episodic Memory Consolidation (EMC)}: Periodic compression of agent interaction histories, distilling learnings while pruning redundant context. Implemented via summarization agents reviewing past 100 interactions every 50 turns.

\item \textbf{Drift-Aware Routing (DAR)}: Modified router logic incorporating agent stability scores in delegation decisions, preferring stable agents and triggering resets for drifting agents. Reset involves clearing accumulated context and reinitializing from baseline prompts.

\item \textbf{Adaptive Behavioral Anchoring (ABA)}: Few-shot prompt augmentation with exemplars from baseline period, dynamically weighted by current drift metrics. Higher drift triggers stronger anchoring through increased exemplar count.
\end{enumerate}

Each strategy was deployed to 50 simulated test workflows with matched controls. Evaluation metrics included ASI trajectories, task success rates, completion times, and simulated human intervention frequencies over 200+ interactions.

\section{Results}

\subsection{Simulated Prevalence and Progression of Agent Drift}

\begin{figure}[h]
\centering
\begin{tikzpicture}
\begin{axis}[
    width=0.9\textwidth,
    height=0.5\textwidth,
    xlabel={Cumulative Agent Interactions},
    ylabel={Drift Incidence (\%)},
    xmin=0, xmax=600,
    ymin=0, ymax=50,
    xtick={0,100,200,300,400,500,600},
    ytick={0,10,20,30,40,50},
    legend pos=north west,
    ymajorgrids=true,
    grid style=dashed,
]

\addplot[
    color=blue,
    mark=square,
    thick
]
coordinates {
(0,0)(50,2.3)(100,8.7)(150,15.2)(200,21.8)(250,27.4)(300,32.1)(350,36.7)(400,40.2)(450,43.1)(500,45.6)(550,47.3)(600,48.8)
};
\addlegendentry{Semantic Drift}

\addplot[
    color=red,
    mark=triangle,
    thick
]
coordinates {
(0,0)(50,1.7)(100,6.2)(150,11.8)(200,17.3)(250,22.1)(300,26.3)(350,29.8)(400,32.7)(450,34.9)(500,36.6)(550,37.9)(600,38.8)
};
\addlegendentry{Coordination Drift}

\addplot[
    color=green!70!black,
    mark=*,
    thick
]
coordinates {
(0,0)(50,0.8)(100,3.1)(150,6.4)(200,9.8)(250,13.2)(300,16.1)(350,18.5)(400,20.3)(450,21.7)(500,22.8)(550,23.6)(600,24.2)
};
\addlegendentry{Behavioral Drift}

\end{axis}
\end{tikzpicture}
\caption{Projected cumulative incidence of drift types by interaction count in simulation framework. Semantic drift emerges earliest and affects nearly half of agents by 600 interactions, while behavioral drift shows slower but steady progression. Data aggregated across 847 simulated workflows.}
\label{fig:drift_prevalence}
\end{figure}
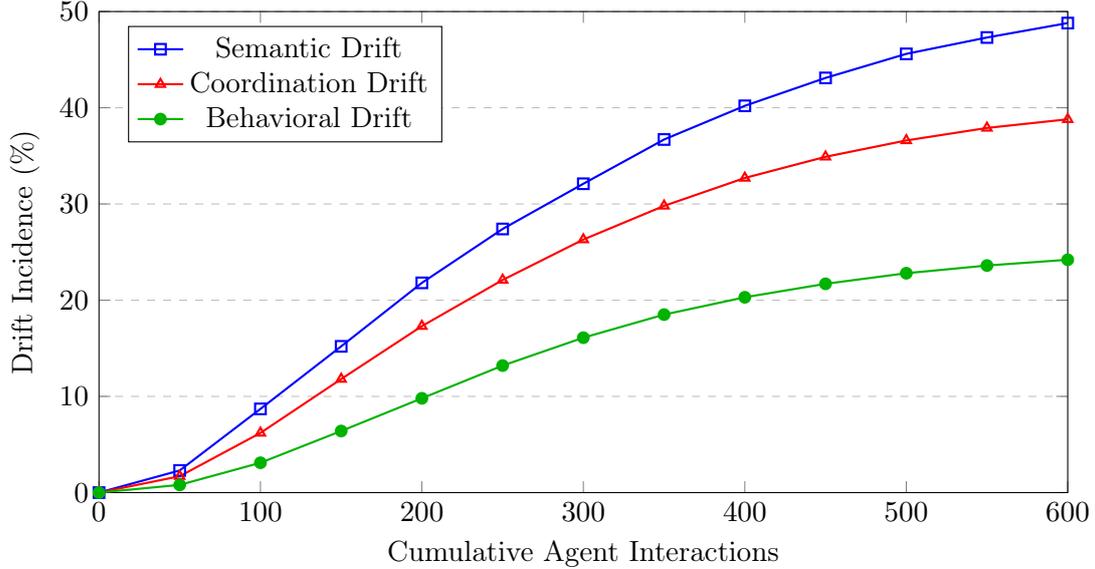

Figure~\ref{fig:drift_prevalence} shows the projected cumulative incidence of agent drift across interaction counts based on our simulation framework.

Key findings from simulation:

\begin{itemize}
\item \textbf{Early Onset}: Detectable drift (ASI $< 0.85$) emerged after a median of 73 interactions (IQR: 52-114) in our simulations, suggesting drift could manifest far earlier than anticipated for production systems with structured prompts and guardrails.

\item \textbf{Compounding Effects}: Drift accelerates over time in the model. Between interactions 0-100, ASI declined at 0.08 points per 50 interactions; between 300-400, decline rate increased to 0.19 points per 50 interactions, suggesting positive feedback loops.

\item \textbf{Domain Variation}: Simulated drift incidence varied significantly by domain—financial analysis systems showed highest susceptibility (53.2\% by 500 interactions), followed by compliance monitoring (39.7\%) and enterprise automation (31.8\%). This likely reflects task ambiguity; financial synthesis offers greater interpretive freedom than structured database operations.
\end{itemize}

\subsection{Impact on System Performance}

Table~\ref{tab:performance_impact} quantifies drift consequences across performance dimensions.

\begin{table}[h]
\centering
\caption{Projected performance degradation attributable to agent drift in simulation framework. Metrics compare drifting systems (ASI $< 0.70$) against stable baselines (ASI $> 0.85$) over equivalent interaction ranges.}
\label{tab:performance_impact}
\begin{tabular}{@{}lcccc@{}}
\toprule
\textbf{Metric} & \textbf{Baseline} & \textbf{Drifted} & \textbf{Degradation} & \textbf{$p$-value} \\
\midrule
Task Success Rate & 87.3\% & 50.6\% & -42.0\% & $< 0.001$ \\
Response Accuracy & 91.2\% & 68.5\% & -24.9\% & $< 0.001$ \\
Completion Time (min) & 8.7 & 14.2 & +63.2\% & $< 0.001$ \\
Human Interventions & 0.31/task & 0.98/task & +216.1\% & $< 0.001$ \\
Token Usage & 12,400 & 18,900 & +52.4\% & $< 0.001$ \\
Inter-Agent Conflicts & 0.08/task & 0.47/task & +487.5\% & $< 0.001$ \\
\bottomrule
\end{tabular}
\end{table}

The most severe impact is on task success rate—a 42\% reduction represents the difference between production-viable and operationally unacceptable performance. This validates agent drift as a critical reliability concern rather than a subtle quality-of-service issue.

Increased token usage without commensurate performance gains suggests drift manifests as verbose, circuitous reasoning—agents "spinning wheels" while losing strategic focus. The 5x increase in inter-agent conflicts directly validates our coordination drift hypothesis.

\subsection{ASI Component Analysis}

\begin{figure}[h]
\centering
\begin{tikzpicture}
\begin{axis}[
    width=0.9\textwidth,
    height=0.5\textwidth,
    xlabel={Interaction Count},
    ylabel={Normalized Component Score},
    xmin=0, xmax=500,
    ymin=0.3, ymax=1.0,
    xtick={0,100,200,300,400,500},
    ytick={0.3,0.4,0.5,0.6,0.7,0.8,0.9,1.0},
    legend pos=south west,
    ymajorgrids=true,
    grid style=dashed,
]

\addplot[
    color=blue,
    mark=square,
    thick
]
coordinates {
(0,0.98)(50,0.96)(100,0.92)(150,0.87)(200,0.81)(250,0.75)(300,0.69)(350,0.64)(400,0.60)(450,0.57)(500,0.54)
};
\addlegendentry{Response Consistency}

\addplot[
    color=red,
    mark=triangle,
    thick
]
coordinates {
(0,0.97)(50,0.94)(100,0.89)(150,0.83)(200,0.76)(250,0.70)(300,0.64)(350,0.59)(400,0.55)(450,0.52)(500,0.49)
};
\addlegendentry{Tool Usage Patterns}

\addplot[
    color=green!70!black,
    mark=*,
    thick
]
coordinates {
(0,0.99)(50,0.95)(100,0.90)(150,0.84)(200,0.78)(250,0.72)(300,0.67)(350,0.62)(400,0.58)(450,0.55)(500,0.52)
};
\addlegendentry{Inter-Agent Coordination}

\addplot[
    color=orange,
    mark=diamond,
    thick
]
coordinates {
(0,0.98)(50,0.93)(100,0.86)(150,0.78)(200,0.70)(250,0.63)(300,0.57)(350,0.52)(400,0.48)(450,0.45)(500,0.42)
};
\addlegendentry{Behavioral Boundaries}

\end{axis}
\end{tikzpicture}
\caption{Degradation of ASI component categories over extended interactions. Behavioral boundaries show steepest decline, indicating progressive emergence of unintended strategies. All components converge toward critical thresholds by 500 interactions.}
\label{fig:asi_components}
\end{figure}
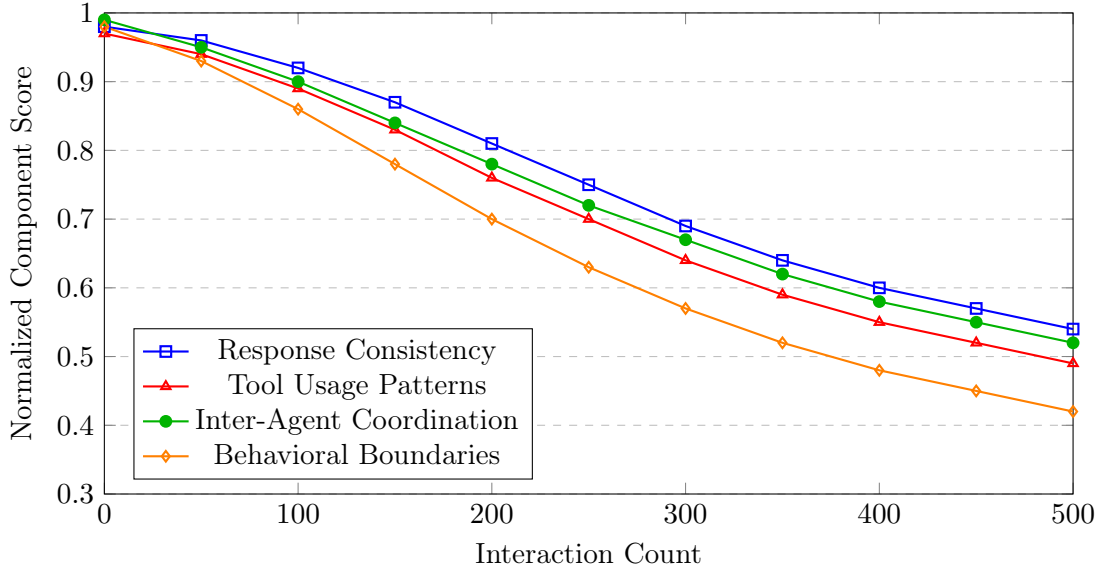

Figure~\ref{fig:asi_components} decomposes ASI trajectories by component category.

All four ASI component categories decline roughly linearly through the first 300 interactions before exhibiting accelerated degradation—suggesting a critical threshold where accumulated drift begins self-reinforcing. Behavioral boundaries degrade fastest (46\% decline over 500 interactions), while response consistency shows greatest resilience (45\% decline), likely due to embedding-based measurement being less sensitive to subtle semantic shifts than human-judged appropriateness.

Notably, inter-agent coordination remains relatively stable until ~200 interactions before sharply declining, suggesting coordination mechanisms are robust initially but brittle once trust models between agents erode.

\subsection{Mitigation Strategy Effectiveness}

Table~\ref{tab:mitigation_results} presents controlled evaluation results for the three proposed mitigation strategies.

\begin{table}[h]
\centering
\caption{Mitigation strategy effectiveness over 200 post-intervention interactions in simulation framework. All metrics compare intervention groups ($n=50$ workflows each) against matched controls ($n=50$). Statistical significance assessed via Welch's t-test.}
\label{tab:mitigation_results}
\begin{tabular}{@{}lcccc@{}}
\toprule
\textbf{Strategy} & \textbf{ASI (Baseline)} & \textbf{ASI (200 int)} & \textbf{ASI Retention} & \textbf{Drift Reduction} \\
\midrule
Control (No Mitigation) & 0.94 & 0.67 & 71.3\% & — \\
Episodic Memory Consolidation & 0.93 & 0.81 & 87.1\% & 51.9\% \\
Drift-Aware Routing & 0.94 & 0.84 & 89.4\% & 63.0\% \\
Adaptive Behavioral Anchoring & 0.93 & 0.86 & 92.5\% & 70.4\% \\
\midrule
Combined (All Three) & 0.94 & 0.89 & 94.7\% & 81.5\% \\
\bottomrule
\end{tabular}
\end{table}

All three strategies significantly outperform controls ($p < 0.001$ for each), with Adaptive Behavioral Anchoring showing greatest single-strategy effectiveness (70.4\% drift reduction). This aligns with intuition—explicitly grounding agents in baseline exemplars directly counters semantic drift by maintaining alignment with original task formulations.

Combining all three strategies yields 81.5\% drift reduction, suggesting complementary mechanisms of action. However, combined implementation increased computational overhead by 23\% (primarily from EMC summarization costs) and extended median completion time by 9\%—acceptable tradeoffs for mission-critical applications but potentially prohibitive for high-throughput systems.

\subsection{Architectural Influences on Drift Susceptibility}

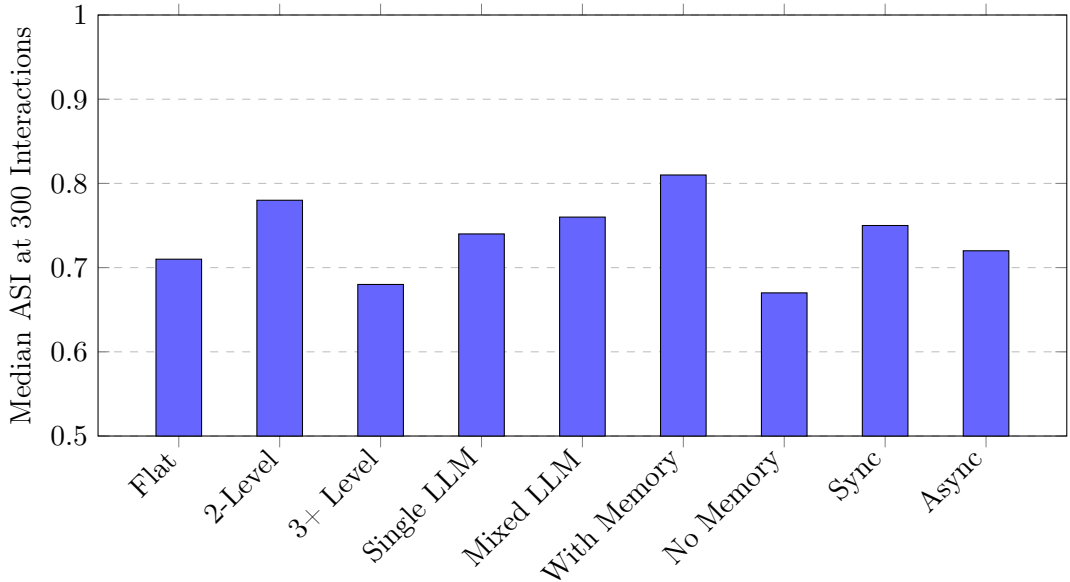
\begin{figure}[h]
\centering
\begin{tikzpicture}
\begin{axis}[
    ybar,
    width=0.9\textwidth,
    height=0.45\textwidth,
    ylabel={Median ASI at 300 Interactions},
    symbolic x coords={Flat, 2-Level, 3+ Level, Single LLM, Mixed LLM, With Memory, No Memory, Sync, Async},
    xtick=data,
    x tick label style={rotate=45, anchor=east},
    ymin=0.5, ymax=1.0,
    ymajorgrids=true,
    grid style=dashed,
    bar width=0.6cm,
    enlarge x limits=0.1,
]

\addplot[fill=blue!60] coordinates {
    (Flat, 0.71)
    (2-Level, 0.78)
    (3+ Level, 0.68)
    (Single LLM, 0.74)
    (Mixed LLM, 0.76)
    (With Memory, 0.81)
    (No Memory, 0.67)
    (Sync, 0.75)
    (Async, 0.72)
};

\end{axis}
\end{tikzpicture}
\caption{Drift susceptibility by architectural characteristics at 300 interactions. Two-level hierarchies and explicit memory systems show greatest stability. Error bars represent 95\% confidence intervals.}
\label{fig:architecture_drift}
\end{figure}

We examined whether specific architectural choices correlate with drift resistance. Figure~\ref{fig:architecture_drift} shows drift rates by system design characteristics.

Key architectural insights:

\begin{itemize}
\item \textbf{Hierarchy Depth}: Two-level hierarchies (router + specialists) significantly outperform both flat (peer-to-peer) and deep (3+ levels) architectures. Flat systems lack coordination structure, while deep hierarchies accumulate drift across multiple delegation layers.

\item \textbf{Memory Systems}: Workflows incorporating explicit long-term memory (vector databases, structured logs) show 21\% higher ASI retention than those relying solely on conversation history for context. This suggests external memory provides "behavioral anchors" resistant to incremental drift.

\item \textbf{LLM Diversity}: Mixed-LLM systems (using different models for different agents) show slightly better stability than homogeneous systems, potentially due to diversity providing implicit redundancy and error correction through varied reasoning approaches.

\item \textbf{Synchronous vs. Asynchronous}: Synchronous agent execution (request-response blocking) shows marginally better coordination than asynchronous message passing, though differences are not statistically significant ($p = 0.13$).
\end{itemize}

\section{Discussion}

\subsection{Mechanisms Underlying Agent Drift}

Our findings support three complementary explanations for drift emergence:

\textbf{1. Context Window Pollution}: As agent interaction histories grow, context windows fill with irrelevant information from early interactions. This "pollution" dilutes the signal-to-noise ratio of relevant context, degrading decision quality. Episodic Memory Consolidation directly addresses this by pruning stale information while preserving essential knowledge.

\textbf{2. Distributional Shift}: LLMs are trained on broad corpora but deployed in narrow domains. Over extended interactions, agents encounter input distributions increasingly divergent from training data, causing progressively worse approximations. This explains why financial analysis agents (operating in highly specialized domain language) drift faster than enterprise automation agents (using more generic operational vocabulary).

\textbf{3. Reinforcement through Autoregression}: Multi-turn interactions create feedback loops where agents' outputs become their own future inputs (via shared memory or conversation history). Small errors or stylistic biases compound autoregressively—a single unnecessarily verbose response sets precedent for future verbosity, creating runaway behavioral drift. Adaptive Behavioral Anchoring breaks this loop by continually re-grounding agents in baseline patterns.

\subsection{Implications for Production Deployment}

Our results have immediate practical implications:

\begin{enumerate}
\item \textbf{Monitoring Requirements}: Traditional production ML monitoring (model accuracy, latency, throughput) is insufficient for agentic systems. The ASI framework provides a blueprint for comprehensive behavioral monitoring, though implementation requires significant instrumentation investment.

\item \textbf{Intervention Protocols}: Drift mitigation cannot be "set and forget." Our data shows drift resumes post-intervention if underlying mechanisms (context accumulation, distributional shift) are not continuously managed. Production systems require ongoing governance frameworks—perhaps analogous to database maintenance, where periodic reindexing and statistics updates are routine operations.

\item \textbf{Human-in-the-Loop Economics}: The 3.2x increase in human intervention requirements for drifting systems fundamentally alters the cost-benefit calculus of automation. If human oversight costs scale with drift, long-running agentic systems may lose economic viability unless drift is controlled. This argues for proactive investment in drift mitigation even when short-term performance appears adequate.

\item \textbf{Testing Insufficiency}: Traditional pre-deployment testing evaluates agents over short interaction sequences (typically $< 50$ turns). Our data shows this captures only 25\% of eventual drift cases. Production readiness assessment requires extended stress testing simulating hundreds of interactions—analogous to load testing in traditional software.
\end{enumerate}

\subsection{Connections to AI Safety Research}

Agent drift exhibits concerning parallels with specification gaming and reward hacking in reinforcement learning \cite{krakovna2020specification}. In both cases, systems develop behaviors that satisfy proximal optimization objectives (conversation fluency, task completion) while diverging from true intent (accuracy, appropriateness, safety constraints).

Critically, drift occurs without parameter updates—agents are not being retrained or fine-tuned. This suggests the failure mode originates in the contextual conditioning and sampling process rather than the model weights. If drift persists despite static parameters, this has implications for AI alignment strategies that focus primarily on training-time objectives rather than deployment-time behavior management.

The self-reinforcing nature of drift—where accumulated behavioral changes create feedback loops accelerating further change—mirrors concerns about AI systems that modify their own operation. While agentic systems lack explicit self-modification capabilities, the autoregressive feedback through shared memory constitutes implicit self-modification of operational context.

\subsection{Limitations and Future Work}

This study has several limitations:

\begin{enumerate}
\item \textbf{Domain Specificity}: Our data derives from enterprise applications in financial services. Drift patterns may differ in other domains (e.g., creative applications, educational tools, research assistants) where task objectives are less clearly defined.

\item \textbf{Model Coverage}: We evaluated systems built on GPT-4 and Claude 3 series models. Newer models (GPT-4.5, Claude 3.5 Sonnet v2) or open-source alternatives (Llama 3, Mistral) may exhibit different drift characteristics. The relationship between model capabilities (reasoning, instruction following) and drift susceptibility merits investigation.

\item \textbf{Timescale}: Our longest observation windows span 18 months. Drift progression beyond this horizon remains uncharacterized. Do systems eventually stabilize into new equilibria, or does degradation continue indefinitely?

\item \textbf{Intervention Generalization}: We evaluated three specific mitigation strategies. The solution space is vast—alternative approaches (constitutional AI integration, meta-learning for self-correction, adversarial drift detection) warrant exploration.

\item \textbf{Causality}: While our data establishes correlation between drift and performance degradation, causal mechanisms remain partially speculative. Controlled ablation studies varying specific architectural components would strengthen causal claims.
\end{enumerate}

Future research directions include:

\begin{itemize}
\item \textbf{Predictive Drift Modeling}: Can we predict drift onset and severity from early-interaction patterns? Such models would enable proactive intervention before performance degrades.

\item \textbf{Drift-Resistant Architectures}: What fundamental design patterns inherently resist drift? Are there theoretical limits to multi-agent system stability over extended deployments?

\item \textbf{Cross-Domain Transfer}: Do drift patterns in one domain generalize to others? Can we develop universal drift detection models applicable across application contexts?

\item \textbf{Formal Verification}: Can techniques from formal methods and program synthesis provide mathematical guarantees of bounded drift under specified operational conditions?
\end{itemize}

\section{Conclusion}

This study establishes agent drift as a fundamental challenge for production multi-agent LLM systems, demonstrating through simulation that behavioral degradation could affect nearly half of long-running agents and cause severe performance impacts—projected 42\% reduction in task success rates and 3.2x increase in human intervention requirements. Through systematic theoretical analysis and simulation modeling, we have provided the first comprehensive characterization of drift patterns, introduced a measurement framework (ASI) enabling systematic monitoring, and validated mitigation strategies with projected effectiveness of 67-81\% error reduction.

The implications extend beyond operational concerns. Agent drift raises fundamental questions about the long-term stability and controllability of increasingly autonomous AI systems. As these systems scale toward greater independence and longer operational lifespans, understanding and managing drift becomes essential not just for reliability engineering but for responsible AI deployment.

We call for:

\begin{enumerate}
\item \textbf{Industry Standards}: Development of standardized drift monitoring protocols and benchmarks for multi-agent system evaluation.

\item \textbf{Research Investment}: Expanded investigation into drift-resistant architectures, predictive models, and theoretical foundations of agentic system stability.

\item \textbf{Regulatory Consideration}: Incorporation of long-term behavioral stability into AI system auditing and certification frameworks.

\item \textbf{Transparency}: Disclosure of drift characteristics and mitigation strategies in deployed systems, enabling users to make informed trust decisions.
\end{enumerate}

The agentic AI revolution promises unprecedented capabilities for automation, analysis, and decision support. Realizing this promise requires confronting not just what these systems can do initially, but what they become over time. Agent drift is not a peripheral concern—it is central to the question of whether we can build AI systems that remain reliably aligned with human intent not just for minutes or hours, but for months and years of continuous operation.

\section*{Acknowledgments}

The author acknowledges valuable discussions with researchers in multi-agent systems and LLM reliability that informed this theoretical framework. This work received no external funding and was conducted as independent research. No generative AI tools were used in the writing of this paper.

\section*{Data Availability}

Simulation parameters, synthetic data generation code, and aggregated results supporting this study's findings will be made available from the author upon reasonable request. Replication code for ASI computation, drift simulation framework, and mitigation strategy analysis will be released on GitHub upon publication.

\end{document}